\documentclass[10pt,twocolumn,letterpaper]{article}

\usepackage{cvpr}              
\usepackage[utf8]{inputenc} 
\usepackage[T1]{fontenc}    
\usepackage{microtype}      
\usepackage{wrapfig}
\usepackage{xspace}
\usepackage{subcaption}
\usepackage{placeins}
\usepackage{bmpsize}
\usepackage{multirow}
\usepackage{adjustbox}
\usepackage{makecell}
\usepackage{arydshln}

\usepackage[symbol]{footmisc} 

\usepackage{url}            
\usepackage{booktabs}       
\usepackage{nicefrac}       
\usepackage{graphicx}
\usepackage{amsfonts,amssymb,amsmath,amsthm}
\usepackage{multirow}
\usepackage{pifont}
\usepackage{algorithm}
\usepackage{algorithmic}
\usepackage[pagebackref,breaklinks,colorlinks,allcolors=cvprblue]{hyperref}
\usepackage[capitalize]{cleveref} 
\definecolor{applegreen}{rgb}{0.55, 0.71, 0.0}

\newcommand{\cmark}{\ding{51}}%
\newcommand{\xmark}{\ding{55}}%
\definecolor{cvprblue}{rgb}{0.21,0.49,0.74}
\definecolor{darkgreen}{RGB}{0,130,0}
\definecolor{darkred}{RGB}{180,0,0}

\def\paperID{12037} 
\def\confName{CVPR}
\def\confYear{2025}

\title{SocialGesture: Delving into Multi-person Gesture Understanding}

\author{
Xu Cao$^{1}$, Pranav Virupaksha$^{1}$, Wenqi Jia$^{1}$, Bolin Lai$^{2}$, Fiona Ryan$^{2}$, Sangmin Lee$^{3}$\thanks{Corresponding author} , James M. Rehg$^{1}$\footnotemark[2] \\
$^1$ University of Illinois Urbana-Champaign \quad $^2$ Georgia Institute of Technology \\
$^{3}$Sungkyunkwan University \\
{\tt\small \{xucao2,pranavv3,wenqij5,jrehg\}@illinois.edu \{bolin.lai,fkryan\}@gatech.edu} \\ {\tt\small sangmin.lee@skku.edu}
}

\begin{document}
\maketitle

\begin{abstract}

Previous research in human gesture recognition has largely overlooked multi-person interactions, which are crucial for understanding the social context of naturally occurring gestures. This limitation in existing datasets presents a significant challenge in aligning human gestures with other modalities like language and speech. To address this issue, we introduce SocialGesture, the first large-scale dataset specifically designed for multi-person gesture analysis. SocialGesture features a diverse range of natural scenarios and supports multiple gesture analysis tasks, including video-based recognition and temporal localization, providing a valuable resource for advancing the study of gesture during complex social interactions. Furthermore, we propose a novel visual question answering (VQA) task to benchmark vision language models' (VLMs) performance on social gesture understanding. Our findings highlight several limitations of current gesture recognition models, offering insights into future directions for improvement in this field. SocialGesture is available at \href{https://huggingface.co/datasets/IrohXu/SocialGesture}{huggingface.co/datasets/IrohXu/SocialGesture}.

\end{abstract}
\section{Introduction}

Long before the development of spoken language in human history, deictic gestures played an important role in human social communication, serving as one of the earliest communication tools for expressing thoughts, emotions, and intentions. In contemporary communication, both verbal (e.g., speech) and non-verbal (e.g., gesture) cues work in concert. While spoken words express direct meanings, understanding the complete social context often requires more than just the words alone. Gestures play an important role in clarifying intention or points of emphasis, and augment spoken communication with additional social context. 
Theoretical analysis suggests that both speech and gesture originate from a shared cognitive system, with gestures representing imagery-driven data and speech representing symbolic data~\cite{bernardis2006speech}. Together, they reflect a unified communicative intent.


\begin{figure}[!t]
    \centering
    \includegraphics[width=1.0\linewidth]{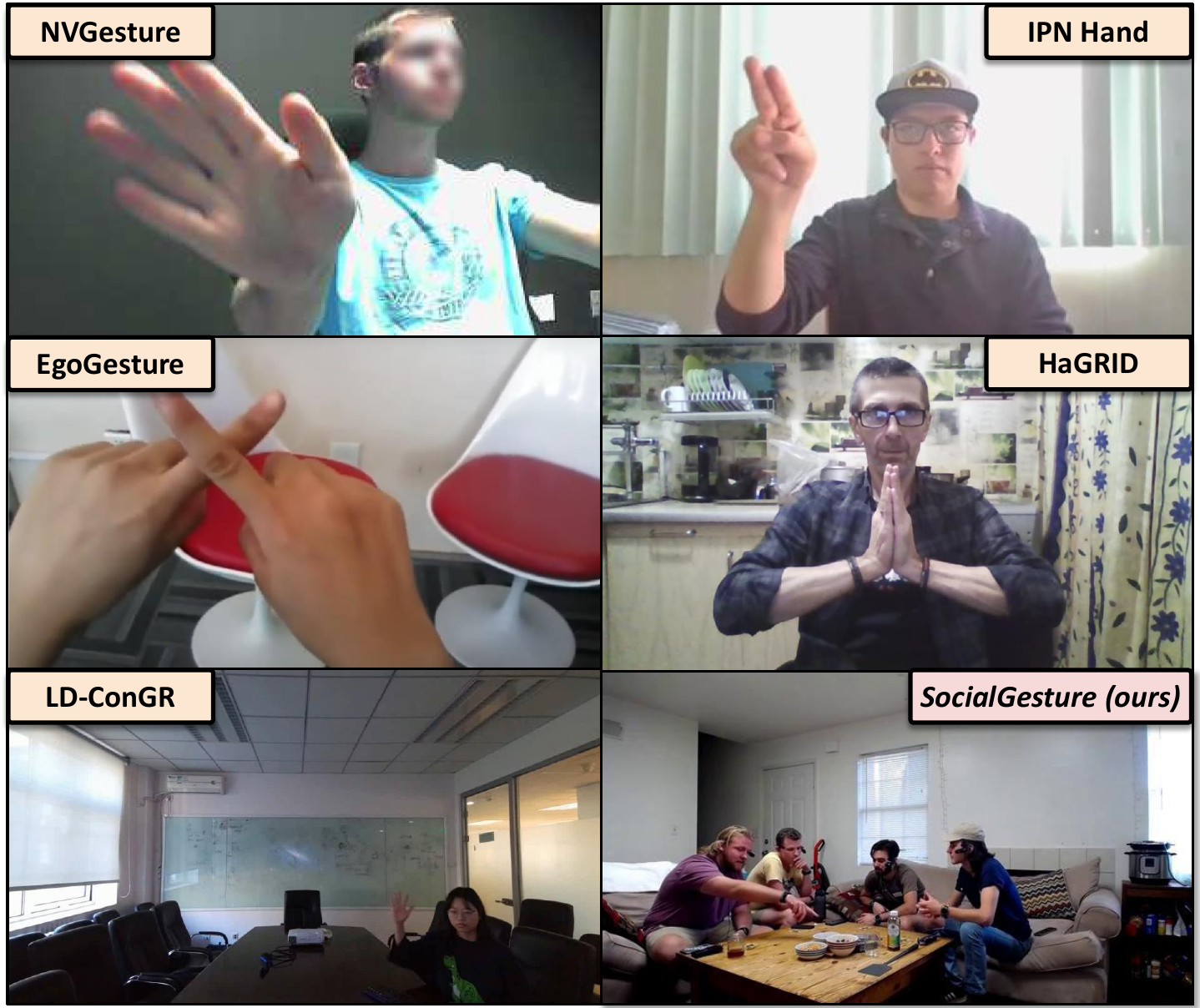}
    \vspace{-0.65cm}
    \caption{Example frames from six gesture datasets. SocialGesture is the only dataset featuring \textit{multi-person interactions} and focusing on natural gestures with meaningful social communication.}
    \label{fig:datasets_comparison}
    \vspace{-0.35cm}
\end{figure}

\begin{table*}[!t]
\renewcommand{\arraystretch}{0.9}
\renewcommand{\tabcolsep}{3.8mm}
\centering
\adjustbox{max width=1.0\linewidth}{
\begin{tabular}{lccccccc}
\toprule
\multirow{2}{*}{\textbf{Dataset}} & \multirow{2}{*}{\textbf{Multi-person}}  & \multirow{2}{*}{\textbf{Video Clips}} & \multirow{2}{*}{\textbf{Instances}} & \multicolumn{4}{c}{\textbf{Label}} \\ \cmidrule(lr){5-8} 
&&&& \textbf{Category} & \textbf{Bbox} & \textbf{Relation} & \textbf{VQA} \\ 
\midrule
Jester~\cite{materzynska2019jester} & {\Large \color{darkred} \xmark}  & 148,092 & 148,092 & \Large \color{darkgreen} \cmark & \Large \color{darkred} \xmark & \Large \color{darkred} \xmark & \Large \color{darkred} \xmark \\
NVGesture~\cite{molchanov2016online} & {\Large \color{darkred} \xmark}  & 1,532 & 1,532 & \Large \color{darkgreen} \cmark & \Large \color{darkgreen} \cmark & \Large \color{darkred} \xmark & \Large \color{darkred} \xmark \\
EgoGesture~\cite{zhang2018egogesture} & {\Large \color{darkred} \xmark}  & 2,081 & 24,161 & \Large \color{darkgreen} \cmark & \Large \color{darkgreen} \cmark & \Large \color{darkred} \xmark & \Large \color{darkred} \xmark \\
ChaLearn ConGD~\cite{wan2016chalearn} & {\Large \color{darkred} \xmark}  & 22,535 & 47,933 & \Large \color{darkgreen} \cmark & \Large \color{darkgreen} \cmark & \Large \color{darkred} \xmark & \Large \color{darkred} \xmark \\
IPN Hand~\cite{benitez2021ipn} & {\Large \color{darkred} \xmark}  & 200 & 4,218 & \Large \color{darkgreen} \cmark & \Large \color{darkgreen} \cmark & \Large \color{darkred} \xmark & \Large \color{darkred} \xmark \\
UAV gesture~\cite{perera2018uav} & {\Large \color{darkred} \xmark}  & 119 & - & \Large \color{darkgreen} \cmark & \Large \color{darkgreen} \cmark & \Large \color{darkred} \xmark & \Large \color{darkred} \xmark \\
iMiGUE~\cite{liu2021imigue} & {\Large \color{darkred} \xmark}  & 359 & 18,499 & \Large \color{darkgreen} \cmark & \Large \color{darkred} \xmark & \Large \color{darkred} \xmark & \Large \color{darkred} \xmark \\
LD-ConGR~\cite{liu2022ld} & {\Large \color{darkred} \xmark}  & 542 & 44,887 & \Large \color{darkgreen} \cmark & \Large \color{darkgreen} \cmark & \Large \color{darkred} \xmark & \Large \color{darkred} \xmark \\
\hline
\bf SocialGesture (Ours) & {\Large \color{darkgreen} \cmark}  & 9,889 & 42,533  & \Large \color{darkgreen} \cmark & \Large \color{darkgreen} \cmark & \Large \color{darkgreen} \cmark & \Large \color{darkgreen} \cmark \\
\bottomrule
\end{tabular}
}
\caption{Comparison of SocialGesture with existing video-based gesture datasets. SocialGesture is the first work that addresses multi-person social interaction cases. Our dataset uniquely provides comprehensive annotations including gesture categories, bounding boxes (humans and objects), gesture relations (subject person and target person), and visual question answering pairs.}
\label{tab:dataset_comparison}
\vspace{-0.2cm}
\end{table*}


Recent advancements in machine learning have driven research on the understanding of human social communication~\cite{zhang2023large,lee2024modeling}, yet most of the latest social research predominantly focuses on linguistic modalities, often overlooking visual social cues like gestures~\cite{lee2024towards}. While there have been machine learning works on human gestures, most of them have focused on gestures for device manipulation or sign languages~\cite{perera2018uav,rastgoo2021sign}, and the problem of analyzing natural social gestures between multiple people has not received significant attention.
A primary reason for this limitation is the lack of gesture data and annotations that capture multi-person interactions and the use of gestures in realistic social contexts. While recent datasets such as HaGRID~\cite{kapitanov2024hagrid} and LD-ConGR~\cite{liu2022ld} address high-resolution and long-distance gesture collection (Figure~\ref{fig:datasets_comparison}), they remain confined to controlled, single-person settings, and do not contain the spontaneous, natural gestures between multiple people that characterize real-life social interactions.

To address this limitation, we introduce SocialGesture, a comprehensive dataset designed specifically for understanding gestures in multi-person social interactions. SocialGesture provides three key advantages over existing datasets:

\vspace{0.05cm}
\begin{enumerate}[label=\textbf{\arabic*.}]
    \item It focuses on videos with multiple people, making it the first multi-person gesture dataset, in contrast to previous datasets which lack human-to-human interaction.
    \item SocialGesture includes four types of comprehensive annotations: (a) Social gesture categories, identifying the specific class of a gesture (i.e., pointing, showing, giving, reaching) within social interactions; (b) Temporal-spatial annotations, identifying the appearance of social gestures temporally and spatially through time stamps and bounding boxes. (c) Interaction annotations, capturing interpersonal social dynamics that result from gestures; and (d) Visual question answering (VQA) annotations, which link gestures to descriptive content, facilitating the alignment of gestures with contextual meaning. All of these annotations offer a rich resource for developing models for social gesture understanding.
    \item SocialGesture comprises a large volume of diverse and naturally occurring social interactions sourced from YouTube and Ego4D~\cite{grauman2022ego4d}, and is the largest multi-person gesture dataset available.

\end{enumerate}
\vspace{0.05cm}


Based on our novel SocialGesture dataset, we introduce diverse gesture-related benchmark tasks from temporal localization and recognition to visual question answering. Our experiments reveal significant challenges in multi-person gesture understanding for current architectures. It is the first step toward building agents that have capabilities to seamlessly understand non-verbal gesture cues in social contexts.

\noindent The major contributions of this paper are as follows.
\vspace{0.05cm}
\begin{itemize}
	\item We introduce SocialGesture, the first dataset focusing on multi-person gesture interactions in natural social settings. Unlike existing datasets that are limited to single-person or controlled environments, our dataset captures the complexity of real-world gesture-based social communication.
	\item We provide comprehensive multi-level annotations including gesture categories, temporal-spatial localization, interaction dynamics, and visual question-answering pairs. This rich annotation scheme enables the development of models for understanding social gestures.
        \item We perform comprehensive experiments across gesture localization and recognition, evaluating a wide breadth of architectures and observe multiple challenging areas for future works. We further benchmark multiple large vision-language models on our proposed diverse Social Visual Question-Answering tasks.
\end{itemize}

\begin{figure*}[t]
\begin{minipage}[b]{1.0\linewidth}
    \centering
    \includegraphics[width=1.0\linewidth]{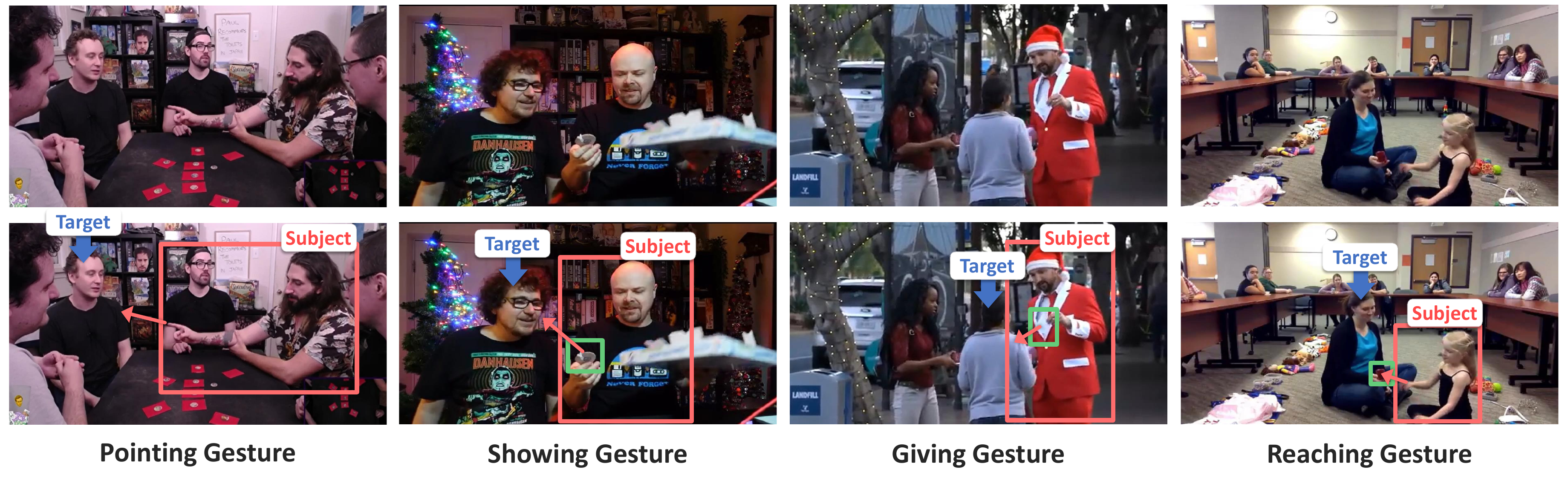}
    \end{minipage}
    \vspace{-0.6cm}
    \caption{Examples of the four deictic gesture categories in SocialGesture with subject-target relationships. From left to right: pointing (directing attention), showing (presenting objects), giving (transfer intention), and reaching (acquisition intention) gestures. Red boxes indicate gesture initiators (subjects) and blue notations indicate targets. }
    \label{fig:gesture_example}
\end{figure*}

\section{Related Works}
\label{sec:rw}

\subsection{Gesture Datasets} As shown in Figure~\ref{fig:datasets_comparison} and Table~\ref{tab:dataset_comparison}, the majority of gesture recognition datasets are video-based, with recent research emphasizing emotional state analysis, increased capture distance for improved model robustness, and egocentric gesture collection. Notable datasets include LD-ConGR, featuring 542 RGB-D videos representing 10 gestures~\cite{liu2022ld}, and HaGRID, which focuses on high-resolution image based gesture recognition across 18 hand gesture classes~\cite{kapitanov2024hagrid}. EgoGesture, with over 2 million frames from 50 subjects, is tailored for egocentric interactions with wearable devices~\cite{zhang2018egogesture}, while iMiGUE~\cite{liu2021imigue} and SMG focus on micro-gesture recognition~\cite{chen2023smg}, containing 359 and 40 videos, respectively. Earlier datasets such as Jester~\cite{materzynska2019jester}, IPN Hand~\cite{benitez2021ipn}, ChaLearn IsoGD~\cite{wan2016chalearn}, and ChaLearn ConGD~\cite{wan2016chalearn} have also made significant contributions, covering tasks like hand segmentation, and isolated or continuous gesture classification. In contrast, we are the first to provide a large-scale multi-person gesture dataset, and we are also the first to provide comprehensive coverage of the four main classes of deictic gestures(defined in \cite{mcneill1992hand}), which play a crucial role in social communication.

\subsection{Social Understanding with Non-verbal Cues} Understanding social interactions has been long studied in the domain of natural language processing \cite{lee2021graph,hazarika2018conversational,ghosal2019dialoguegcn,sacks1974simplest,stolcke2000dialogue,lai2023werewolf}. However, the inherent multimodal nature of social interactions ensures that non-verbal elements (such as gestures, gaze and emotions) play an important role in communication~\cite{lee2024towards}, which are understudied in previous research \cite{yang2024socially}. Recently, several investigations have been conducted on these implicit non-verbal cues. Liu \etal \cite{liu2022ld} build a dataset to identify the gestures of single person from a long distance. Unfortunately, they only focus on isolated gestures of a single person, and fail to contextualize gesture recognition in a real-world social scenario. Non-verbal social understanding of multiple people demands more investigation~\cite{jia2024audio,jahangard2024jrdb,ghosh2024mrac,peng2024tong}. To this end, we make the first attempt to model multi-party social interactions by defining various social gesture understanding tasks, such as social gesture detection, recognition, and VQA, based on our dataset.


\subsection{Foundation Models for Video Understanding} Video foundation models are designed to tackle a wide range of understanding and analysis tasks, including classification, temporal action localization, question answering, captioning, and more. Early approaches to video understanding relied on CNN blocks as visual encoders for video frames, connected to a prediction head or decoder~\cite{feichtenhofer2019slowfast,carreira2017quo}. With the emergence of transformers, new architectures like MViTv2-B~\cite{li2022mvitv2}, VideoSwin~\cite{liu2022video}, UniFormerV2~\cite{li2023uniformerv2}, and VideoMAE-2~\cite{wang2023videomae} tokenize their visual inputs, enabling unified integration with other modalities~\cite{wang2024internvideo2}. Building on this, recently released video-based VLMs like Qwen2-VL~\cite{wang2024qwen2}, LLaVA-NEXT~\cite{liu2024llava}, and InternVL-2.5~\cite{chen2024expanding} leverage pretrained visual encoders to extract rich representations. These models align visual tokens with textual tokens, enabling a deeper multimodal understanding that bridges the gap between video and language tasks. We provide the first investigation of the effectiveness of these models in understanding multi-party social gestures via a VQA task.

\begin{table*}[t]
\renewcommand{\arraystretch}{1.2}
\renewcommand{\tabcolsep}{1.5mm}
\centering
\adjustbox{max width=\linewidth}{
\begin{tabular}{lccccc}
\toprule
\textbf{Video Content} & \textbf{Source} & \textbf{Proportion} & \textbf{Total Length} & \textbf{\# of People} & \textbf{Description} \\ 
\midrule
\multirow{2}{*}{Group social games} & YouTube & 44.51\% & 896 min & >=3  & \multirow{2}{*}{A group of people playing social games such as one night werewolf} \\
& Ego4D & 21.91\% & 441 min & >=4  &  \\
Variety entertainment & YouTube & 22.31\% & 449 min & >=2 & People performing activities such as pranks, challenges, and interviews \\
Educational play & YouTube & 2.53\% & 51 min & >=2 &  Children engaged in educational activities with adult guidance \\
Product reviews & YouTube & 2.53\%& 51 min & >=2  &  People recommending products such as advent calendars \\
Party \& dinner & YouTube & 3.63\% & 73 min & >=7 & A group of people having party \& dinner together \\
Group cooking & YouTube & 2.58\% & 52 min & >=3 & People engaged in cooking activities together \\
\bottomrule
\end{tabular}
}
\vspace{-0.3cm}
\caption{SocialGesture contains diverse videos including different multi-person social interactions from YouTube and Ego4D~\cite{grauman2022ego4d}.}
\label{tab:dataset}
\vspace{-0.4cm}
\end{table*}

\section{SocialGesture}

\subsection{Motivation}

Social gestures comprise a fundamental set of human behavioral practices that facilitate interpersonal communication through hand and body movements in social contexts, distinct from sign language. According to \cite{mcneill1992hand}, gestures can be classified into four primary categories: deictic, beat, iconic, and metaphoric. Deictic gestures further subdivide into pointing, showing, giving, and reaching. Beat gestures are rhythmic movements that emphasize certain words and phrases during speech. Iconic gestures visually represent concrete objects or actions (e.g., making a circle with fingers), while metaphoric gestures symbolize abstract concepts (e.g., making a circular motion to indicate repetition).

Among these categories, deictic gestures deserve particular attention due to their prevalence and significance in social interactions. While iconic and metaphoric gestures appear in specific contexts and beat gestures primarily serve prosodic functions, deictic gestures play a crucial role in establishing joint attention and facilitating object-mediated social interaction. Therefore, our dataset focuses on four fundamental deictic gestures (see Figure~\ref{fig:gesture_example}):

\begin{itemize}

    \item \textbf{Pointing Gesture} involves directing attention to a specific entity in the social scene through finger extension. While commonly executed with an extended index finger and arm, these specific forms are not mandatory requirements. The crucial identifying factor is the communicative intent to guide others' attention to a particular target, whether person or object. This distinguishes pointing from incidental hand movements or other gesture types, particularly reaching gestures where the primary intent differs.
   
    \item \textbf{Showing Gesture} involves presenting an object for others to see it by manipulating or orienting it towards observers. Unlike pointing, the object itself is the focus of attention, and it does not need to be fully supported by the subject, such as when tilting or sliding an object to make its contents visible. The key criterion for a showing gesture is that the object is being presented for visual inspection, regardless of whether the target person actually looks at it or where the subject is directing their gaze.

    \item \textbf{Giving Gesture} involves manipulating an object with the intention of transferring it to another person’s possession. It differs from pointing because the object originates with the subject, and from showing because the subject is inviting the other person to take possession of the object. The key is the subject's intention to transfer the object, regardless of whether the recipient actually takes it.

    \item \textbf{Reaching Gesture} manifests as a hand extension toward an object, expressing desire for possession or requesting transfer. It is characterized by full finger extension and often includes pre-grasp configuration and associated body movements like leaning forward. Reaching differs from showing and giving because the object is not in the subject’s possession, and it differs from pointing because the subject is expressing a desire to possess the object, not just to draw attention to it. The key aspect of reaching is the intent to obtain the object, regardless of whether the subject can physically retrieve it.

\end{itemize}


\subsection{Data Acquisition}

\paragraph{Data collection.} 

We collect raw video data for SocialGesture from various sources, including YouTube channels and Ego4D. Table \ref{tab:dataset} represents the overall composition of the dataset. Video selection is based on four key criteria: (1) Video quality – we manually check each video to ensure high resolution; (2) Number of people – each video is carefully selected to include between two and ten people, ensuring clear visibility of gestures; (3) Video length – the total length of selected videos ranges from 2 to 30 minutes, with each gesture lasting between 1 to 15 seconds; and (4) Scene diversity – to enhance generalization, we select videos featuring diverse races, genders, and ages.

\vspace{-0.4cm}
\paragraph{Pre-processing.}

For videos longer than 10 minutes, we cut them into video clips of 5 minutes in length. All videos are then converted to 720p ($1280\times720$) with 30 FPS in the initial curation and are then mapped to 360p ($640\times360$) with 5 FPS for further annotation.  

\vspace{-0.4cm}
\paragraph{Gesture annotation.}


Each gesture in the SocialGesture dataset is defined as a sequence of video frames where an individual performs an arm or hand movement corresponding to one of four specified deictic gestures: pointing, showing, giving, and reaching. The annotations capture both temporal and spatial aspects of these social interactions through a comprehensive multi-level annotation framework. For temporal annotation, we identify the complete sequence of frames containing a gesture, from initial movement through completion. Within this sequence, we designate a key frame that best captures the defining moment of the gesture - typically when the gesture reaches its most distinctive configuration. This key frame serves as an anchor point for additional annotations. Spatial annotation includes precise bounding box coordinates [x1, y1, x2, y2] for both the gesture initiator and the target. The initiator bounding box encompasses the person performing the gesture, while the target bounding box identifies either the person or object toward which the gesture is directed. These spatial annotations enable analysis of the geometric relationships between interaction participants. Our annotation framework also includes natural language descriptions of each gesture's social context, capturing the broader interaction dynamics beyond pure spatiotemporal coordinates.

\subsection{Benchmark Tasks}

\subsubsection{Social Gesture Temporal Localization}

\begin{figure}[!t]
    \centering
    \includegraphics[width=1.0\linewidth]{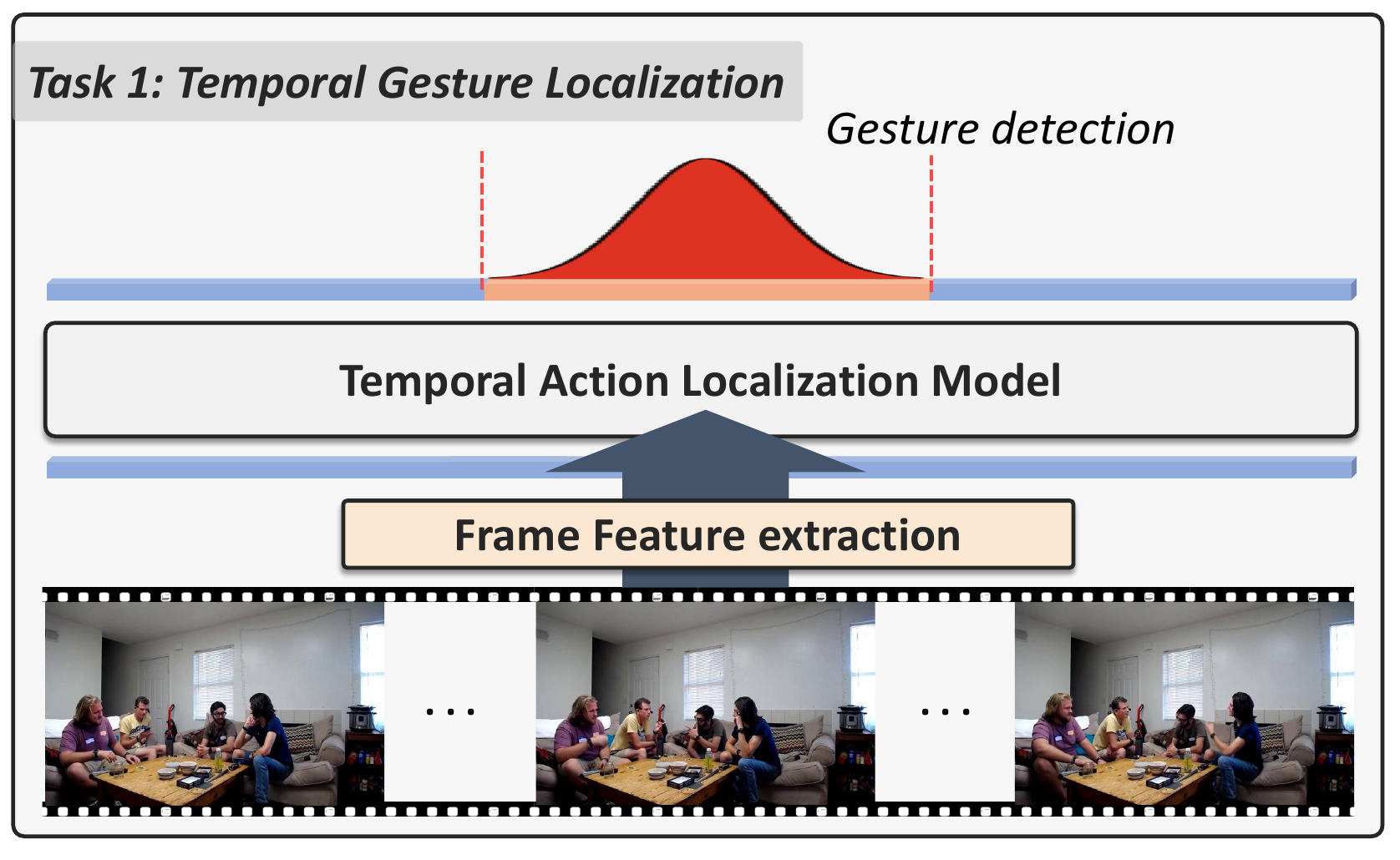}
    \vspace{-0.7cm}
    \caption{Temporal gesture localization task for social gestures}
    \label{fig:localization_task}
    \vspace{-0.4cm}
\end{figure}

Temporal action localization is a challenging task that requires not only identifying the temporal intervals of all detected gesture instances but also estimating corresponding confidence values. Unlike standard classification tasks, this process involves an end-to-end approach in which classification is integrated within temporal localization, adding considerable complexity. For example, even if a gesture is accurately localized in time, it is treated as a false detection if assigned an incorrect class label.

Figure~\ref{fig:localization_task} illustrates a general approach to temporal action localization. Each video is initially divided into multiple overlapping video clips, denoted as $X = [ X_0, X_1, \dots, X_K ]$, where $X_k \in \mathbb{R}^{T \times H \times W \times C}$. Feature extractors such as I3D or VideoMAE are then employed to extract multiscale feature representations, denoted as $Z_k \in \mathbb{R}^{D}$, for each video clip. These features are subsequently concatenated into a single representation, $Z \in \mathbb{R}^{T \times D}$. The objective of the temporal action localization model is to generate sequence labels for each video clip.

This task is also benchmarked by several well-known datasets, including THUMOS14~\cite{idrees2017thumos} and ActivityNet-1.3~\cite{caba2015activitynet}. However, our task presents additional challenges. In SocialGesture videos, multiple individuals are often present simultaneously, and gestures tend to be small, subtle portions of larger actions, making it  uniquely challenging to detect and classify gestures accurately.

\subsubsection{Social Gesture Recognition}

Given the complexity of social gesture localization, we also divide long videos into shorter clips of 2-5 seconds and introduce the social gesture recognition task. This task can be approached in multiple ways. The first approach is isolated gesture recognition, where a long video is segmented into discrete gesture clips. Video action recognition models are then trained to classify the gestures in each clip independently. Alternatively, gestures can be identified continuously throughout the entire video, from beginning to end. Although most action localization methods we discussed in the previous section can typically handle this task, most existing video feature extraction pre-processing methods struggle with consistent multi-person action localization. Therefore, sliding windows remain a common strategy for this purpose: the video is processed in overlapping segments of a predefined window size, with each window analyzed by the model to calculate gesture confidence scores.


We divide recognition into two subtasks (See Figure~\ref{fig:recognition_task}):

\begin{itemize} \item \textbf{Task 2-1: Gesture vs Non-gesture Classification} This task aims to distinguish between social gesture and non-gesture video clips. The input is a short video clip $X \in \mathbb{R}^{T \times H \times W \times C}$, where $T < 16$, and the output is a label $\hat{Y}$ indicating whether a gesture is present.

\begin{figure}[!t]
    \centering
    \includegraphics[width=1.0\linewidth]{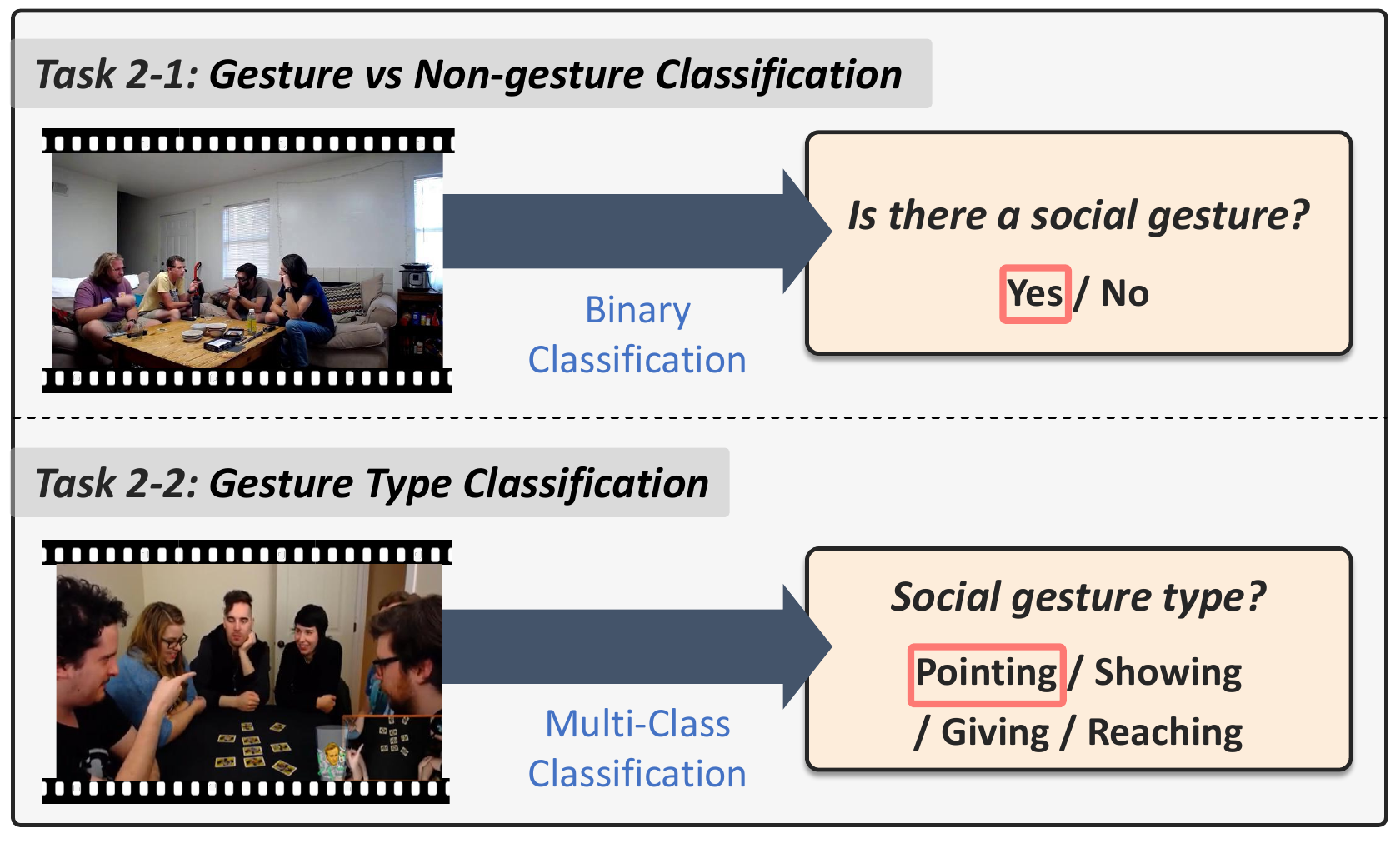}
    \vspace{-0.7cm}
    \caption{Gesture recognition tasks for social gestures}
    \label{fig:recognition_task}
    \vspace{-0.3cm}
\end{figure}

\item \textbf{Task 2-2: Gesture Type Classification.} This task focuses on classifying video clips containing one of four social gestures: pointing, showing, giving, or reaching. The input and output structures are consistent with those used in the binary classification task. We further explore whether using cropped bounding boxes for each individual within a scene improves the accuracy of social gesture classification. The input is a short video clip cropped by the maximum region of the  initiator of a social gesture and the output is a label $\hat{Y}$ indicating four types of social gesture. 

\end{itemize}

For each task, we also strive to sample the training set to achieve a balanced label distribution. This adjustment is necessary because pointing gestures are significantly more frequent than showing, giving, and reaching in real-world communication contexts.

\subsubsection{Social Gesture Visual Question Answering}
\label{subsec:socialgesturevqa}

\begin{figure}[!t]
    \centering
    \includegraphics[width=1.0\linewidth]{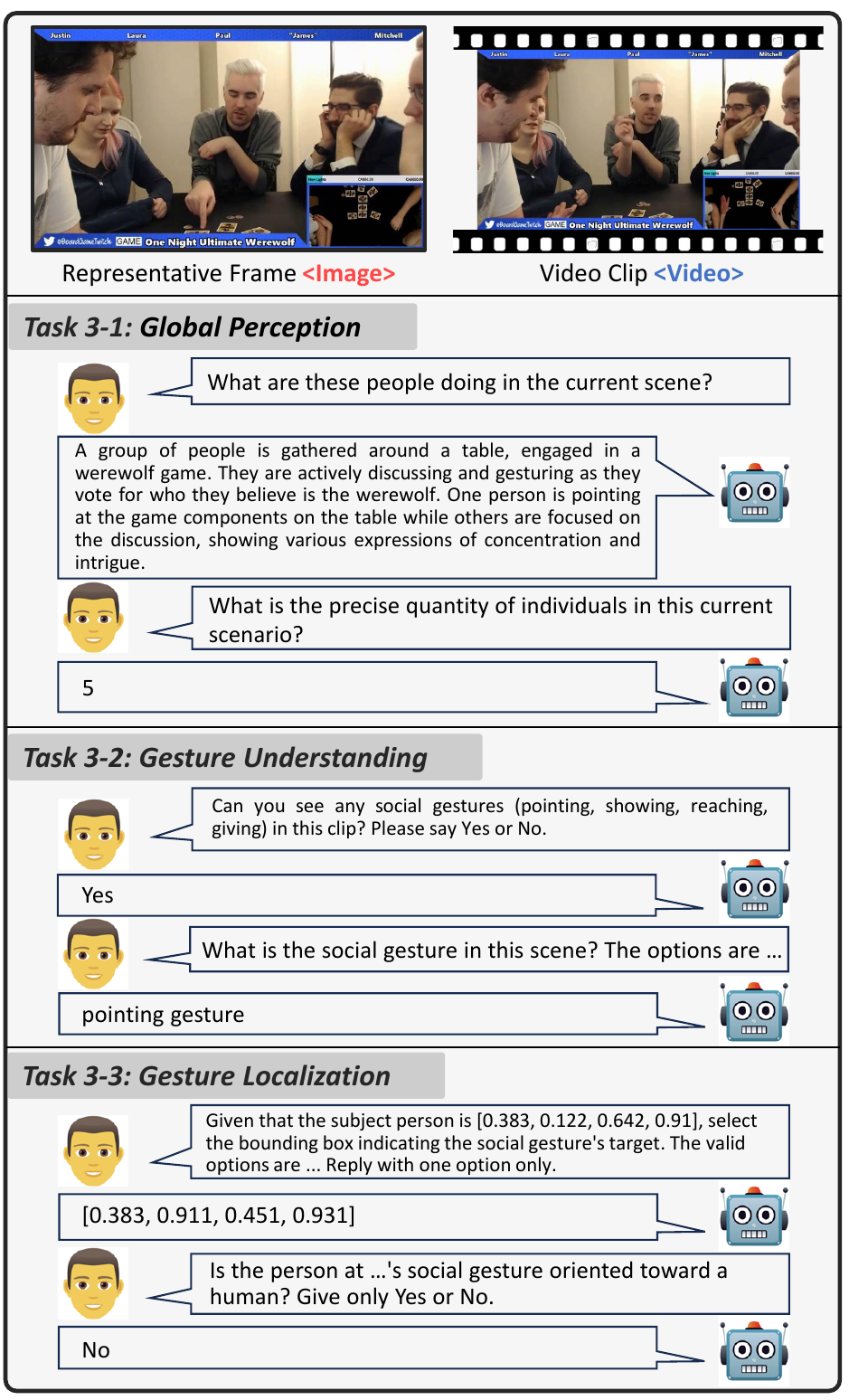}
    \vspace{-0.7cm}
    \caption{The question-answer pairs of SocialGesture. We omit the options of each question in the figure. The bounding box defined by [top-left x, top-left y, bottom-right x, bottom-right y])]. The definition will be provided together with the system prompts.}
    \label{fig:vqa_sample}
    \vspace{-0.3cm}
\end{figure}


Prior to our work, there has been very limited research on designing video captioning and Visual Question Answering (VQA) specifically for human gesture datasets~\cite{tanada2024pointing}. This gap may be due to the lack of multi-person interactions in existing video datasets, reducing the necessity for enhanced alignment between the gesture and language modalities. To fill this gap, we introduce the first VQA task focused on human gestures (Figure~\ref{fig:vqa_sample}). Our VQA can be divided into three subtasks: (1) Global Perception, (2) Gesture Understanding, and (3) Gesture Localization. 

The Global Perception task serves as a foundational test to assess whether VLMs can comprehend basic events in short video clips, including scene descriptions and counting the number of humans present. We generate data for this task through a semi-automated QA generation pipeline based on GPT-4o~\cite{achiam2023gpt}, similar to the instruction-tuning data pair generation in LLaVA~\cite{liu2024visual}. Initially, we manually create short descriptions for each video series, including the social background. We prompt these descriptions with the key frame of the video clip into the GPT-4o pipeline to generate QA pairs for human counting and scene description. Each QA pair is then reviewed by a human annotator. Figure~\ref{fig:vqa_sample} shows the example of global perception (Task 3-1). Human counting is used as one VQA subtask in the VLM benchmarking experiments.

For Gesture Understanding, we utilize gesture classification annotations to create questions related to social gesture classification (Figure~\ref{fig:vqa_sample}, Task 3-2). This results in two types of questions: (1) detecting whether a social gesture occurs, and (2) classifying between the four types of social gestures.

Gesture Localization leverages bounding boxes from our annotations to generate questions that test the model’s spatial and temporal comprehension. These questions include providing the bounding box localization of the initiator, then determining whether the gesture’s target is a human, and locating the target. We present an example of gesture localization (Task 3-3) in Figure~\ref{fig:vqa_sample}. These questions are designed to challenge temporal-spatial reasoning in a multi-person setting, requiring analysis to produce the correct responses.

\section{Experiments}

\subsection{Experimental Setup}

We split the raw videos (long video) into training set (293 videos) and test set (79 videos), and then cut each long video into video clips. The 9,889 video clips used in video action recognition and social gesture VQA tasks come from the same train-test split in the raw videos. We also cut another 4,304 video clips from the raw videos as the non-gesture class. All experiments are conducted with 2 NVIDIA L40S GPUs and 2 H100 GPUs to make sure the batch size is 16 and learning rate is 5e-4 for all models. For all methods in Table~\ref{tab:local_performance}, \ref{tab:recog_binary_performance}, \ref{tab:recog_multiclass_performance}, and \ref{tab:recog_multiclass_performance_bbox}, we apply the same data augmentation strategies used in \cite{touvron2021training,li2023uniformerv2}. To solve the class imbalanced issue, we also adjust the size of training set for each task.

\begin{table}[!t]
\renewcommand{\tabcolsep}{1mm}
\centering
\adjustbox{max width=\linewidth}{
\begin{tabular}{lcccccc}
\toprule
\textbf{Feature} & \textbf{0.3} & \textbf{0.4} & \textbf{0.5} & \textbf{0.6} & \textbf{0.7} & \textbf{Avg mAP} \\ 
\midrule
I3D~\cite{carreira2017quo} & 24.85 & 16.31 & 9.31 & 2.22 & 0.96 & 10.73 \\
R(2+1)D~\cite{tran2018closer} & 14.38 & 10.25 & 7.23 & 2.81 & 1.77 & 7.29 \\
VideoMAEV2~\cite{wang2023videomae} & \textbf{27.23} & \textbf{25.05} & \textbf{13.33} & \textbf{5.28} & \textbf{2.76} & \textbf{14.73} \\
\bottomrule
\end{tabular}
}
\vspace{-0.1cm}
\caption{Temporal action localization for four social gestures by different feature extractors with ActionFormer~\cite{zhang2022actionformer}.}
\label{tab:local_performance}
\end{table}

\begin{table}[!t]
\renewcommand{\tabcolsep}{2.4mm}
\centering
\adjustbox{max width=\linewidth}{
\begin{tabular}{lccccccc}
\toprule
\textbf{Stride} & \textbf{0.3} & \textbf{0.4} & \textbf{0.5} & \textbf{0.6} & \textbf{0.7} & \textbf{Avg mAP} \\ 
\midrule
16 & 11.09 & 8.60 & 5.52 & 2.67 & 1.82 & 5.94\\
8 & 24.85 & 16.31 & 9.31 & 2.22 & 0.96 & 10.73 \\
4 & \textbf{31.64} & \textbf{29.13} & \textbf{19.30} & \textbf{13.77} & \textbf{2.11} & \textbf{19.19} \\
\bottomrule
\end{tabular}
}
\vspace{-0.1cm}
\caption{Explore the influence of stride for I3D and ActionFormer.}
\label{tab:local_performance_ablation}
\vspace{-0.1cm}
\end{table}

\subsection{Social Gesture Temporal Localization}
\vspace{-0.1cm}
\paragraph{Baselines.}
We used two-stream I3D~\cite{carreira2017quo}, R(2+1)D~\cite{tran2018closer}, and VideoMAEV2~\cite{wang2023videomae} with the same sliding window of 16 frames and different stride sizes for feature extraction. Then, these video features are used as input for ActionFormer~\cite{zhang2022actionformer} during model training. We used mAP@[0.3:0.1:0.7] as the main evaluation metric and also reported the average mAP.

\vspace{-0.43cm}
\paragraph{Results and Findings.}
We observe that almost all feature extractors underperform on this task in Table~\ref{tab:local_performance}. Notably, despite the strong performance of features from VideoMAE V2~\cite{wang2023videomae} in other action localization datasets, such as THUMOS 2014~\cite{idrees2017thumos} and ActivityNet~\cite{caba2015activitynet}, they only achieve an average mAP of 14.73 for social gesture recognition. To investigate the effect of the stride size in the sliding window on the average mAP, we conducted an ablation study in Table~\ref{tab:local_performance_ablation} using different stride sizes of the I3D feature in SocialGesture. We found that reducing the stride from 16 to 4 led to an increase in average mAP; however, the overall results remained insufficient. This shortfall can be attributed to the fact that all feature extractors were pretrained on datasets that lack multi-person interactions, causing the features to be poorly aligned with the specific requirements of our task.

\subsection{Social Gesture Recognition}

\textbf{Baselines.} 
Our baselines for social gesture recognition includes different CNN-based video models such as TSN~\cite{wang2016temporal}, TANet~\cite{liu2021tam}, SlowFast~\cite{feichtenhofer2019slowfast}, and Vision Transformer based models such as TimeSformer~\cite{bertasius2021space}, MViTv2~\cite{li2022mvitv2}, VideoSwin~\cite{liu2022video}, UniFormerV2~\cite{li2023uniformerv2}. All these models are pretrained with Kinetics-400~\cite{kay2017kinetics} dataset or using CLIP encoder~\cite{radford2021learning}. We use Accuracy for gesture and non-gesture binary classification in Table~\ref{tab:recog_binary_performance} and Top 1 Accuracy for four social gesture classification in Table~\ref{tab:recog_multiclass_performance}, and \ref{tab:recog_multiclass_performance_bbox}.


\begin{table}[t]
\centering
\adjustbox{max width=\linewidth}{
\begin{tabular}{lccc}
\toprule
\textbf{Model} & \textbf{Pretrain} & \textbf{Param} & \textbf{Acc (\%)} \\ 
\midrule
TSN-R50~\cite{wang2016temporal} & Kinetics-400~\cite{kay2017kinetics} & 24M & 78.77 \\
TANet-R50~\cite{liu2021tam} & Kinetics-400~\cite{kay2017kinetics} & 26M & 71.22 \\
SlowFast-R50~\cite{feichtenhofer2019slowfast} & Kinetics-400~\cite{kay2017kinetics} & 35M & 80.82 \\
SlowFast-R101~\cite{feichtenhofer2019slowfast} & Kinetics-400~\cite{kay2017kinetics} & 63M & 79.59 \\
TimeSformer-L~\cite{bertasius2021space} & Kinetics-400~\cite{kay2017kinetics} & 121M & 78.71 \\
MViTv2-B~\cite{li2022mvitv2} & Kinetics-400~\cite{kay2017kinetics} & 51M & 83.29 \\
VideoSwin-B~\cite{liu2022video} & Kinetics-400~\cite{kay2017kinetics} & 88M & 81.70 \\
VideoSwin-L~\cite{liu2022video} & Kinetics-400~\cite{kay2017kinetics} & 200M & 83.44 \\
UniFormerV2-B/16~\cite{li2023uniformerv2}  &  CLIP~\cite{radford2021learning} & 115M & \textbf{84.43} \\
UniFormerV2-L/14~\cite{li2023uniformerv2}  &  CLIP~\cite{radford2021learning} & 354M & 80.93 \\
\bottomrule
\end{tabular}
}
\vspace{-0.1cm}
\caption{Experiments on binary classification for social gesture and non-gesture.}
\label{tab:recog_binary_performance}
\vspace{-0.2cm}
\end{table}

\vspace{0.1cm}
\noindent \textbf{Results and Findings.} 
The results in Table~\ref{tab:recog_binary_performance} indicate that most existing action recognition models can differentiate social gestures and non-gesture from short video clips. Despite fine-tuning these models using over 10,000 video clips for the binary classification task, their performance remains suboptimal. The SOTA baseline, UniFormerV2-B/16, achieves an accuracy of only 84.43\%. Notably, Transformer-based models generally outperform CNN-based models.

Table~\ref{tab:recog_multiclass_performance} presents results for the four-class social gesture classification task, revealing that all models struggle to effectively classify these gestures—despite the simplicity of the task for human annotators. The highest performance comes from VideoSwim-L, achieving only 56.18\%. Given that individual gestures can be challenging to recognize in multi-person environments, we conducted further experiments, as shown in Table~\ref{tab:recog_multiclass_performance_bbox}. Specifically, we extracted the region corresponding to each individual using ground truth bounding boxes and fine-tuned the same models with these per-person inputs. 
UniFormerV2-B/16 achieves the best performance of 64.72\% top 1 accuracy. This may be because social gestures inherently involve fine-grained and subtle movements. Consequently, recognizing such gestures is difficult from both the extracted per-person region and the full-frame perspective. Addressing the challenge of social gesture understanding requires advanced models capable of disentangling the relationship between the subject and the target person.

\begin{table}[t]
\centering
\adjustbox{max width=\linewidth}{
\begin{tabular}{lcccc}
\toprule
\textbf{Model} & \textbf{Pretrain} & \textbf{Param} & \textbf{Top1 Acc (\%)} \\ 
\midrule
TSN-R50~\cite{wang2016temporal} & Kinetics-400~\cite{kay2017kinetics} & 24M & 54.83 \\
TANet-R50~\cite{liu2021tam} & Kinetics-400~\cite{kay2017kinetics} & 26M & 45.39 \\
SlowFast-R50~\cite{feichtenhofer2019slowfast} & Kinetics-400~\cite{kay2017kinetics} & 35M & 45.17 \\
SlowFast-R101~\cite{feichtenhofer2019slowfast} & Kinetics-400~\cite{kay2017kinetics} & 63M & 46.07 \\
TimeSformer-L~\cite{bertasius2021space} & Kinetics-400~\cite{kay2017kinetics} & 121M & 53.03 \\
MViTv2-B~\cite{li2022mvitv2} & Kinetics-400~\cite{kay2017kinetics} & 51M & 37.98 \\
VideoSwin-B~\cite{liu2022video} & Kinetics-400~\cite{kay2017kinetics} & 88M & 53.93 \\
VideoSwin-L~\cite{liu2022video} & Kinetics-400~\cite{kay2017kinetics} & 200M & \textbf{56.18}  \\
UniFormerV2-B/16~\cite{li2023uniformerv2}  &  CLIP~\cite{radford2021learning} & 115M & 55.51 \\
UniFormerV2-L/14~\cite{li2023uniformerv2}  &  CLIP~\cite{radford2021learning} & 354M & 50.34 \\
\bottomrule
\end{tabular}
}
\vspace{-0.1cm}
\caption{Experiments on classification for four social gestures.}
\label{tab:recog_multiclass_performance}
\end{table}

\begin{table}[!t]
\centering
\adjustbox{max width=\linewidth}{
\begin{tabular}{lccc}
\toprule

\textbf{Model} & \textbf{Pretrain} & \textbf{Param} & \textbf{Top1 Acc (\%)} \\ 
\midrule
TSN-R50~\cite{wang2016temporal} & Kinetics-400~\cite{kay2017kinetics} & 24M & 58.43  \\
TANet-R50~\cite{liu2021tam} & Kinetics-400~\cite{kay2017kinetics} & 26M & 46.29 \\
SlowFast-R50~\cite{feichtenhofer2019slowfast} & Kinetics-400~\cite{kay2017kinetics} & 35M & 45.17  \\
SlowFast-R101~\cite{feichtenhofer2019slowfast} & Kinetics-400~\cite{kay2017kinetics} & 63M & 42.92  \\
TimeSformer-L~\cite{bertasius2021space} & Kinetics-400~\cite{kay2017kinetics} & 121M & 63.60  \\
MViTv2-B~\cite{li2022mvitv2} & Kinetics-400~\cite{kay2017kinetics} & 51M & 35.28 \\
VideoSwin-B~\cite{liu2022video} & Kinetics-400~\cite{kay2017kinetics} & 88M &  60.00 \\
VideoSwin-L~\cite{liu2022video} & Kinetics-400~\cite{kay2017kinetics} & 200M & 63.60 \\
UniFormerV2-B/16~\cite{li2023uniformerv2}  &  CLIP~\cite{radford2021learning} & 115M & \textbf{64.72} \\
UniFormerV2-L/14~\cite{li2023uniformerv2}  &  CLIP~\cite{radford2021learning} & 354M & 60.45 \\

\bottomrule
\end{tabular}
}
\vspace{-0.1cm}
\caption{Experiments on classification for four social gestures after extract each subject person's bounding box.}
\label{tab:recog_multiclass_performance_bbox}
\vspace{-0.3cm}
\end{table}

\begin{table*}[!t]
\renewcommand{\arraystretch}{1.1}
\renewcommand{\tabcolsep}{1.5mm}
\centering
\adjustbox{max width=1.0\linewidth}{
\begin{tabular}{lccccccc}
\toprule
\multirow{2}{*}{\textbf{Model}} & \multirow{2}{*}{\textbf{ Param}}  & \textbf{Global Perception (\%)} & \multicolumn{2}{c}{\textbf{Gesture Understanding (\%)}} & \multicolumn{3}{c}{\textbf{Gesture Localization (\%)}} \\ 
\cmidrule(lr){3-3} \cmidrule(lr){4-5} \cmidrule(lr){6-8}
&  & HumanCount@Acc & GestureDet@Acc & GestureClass@Acc & TargetLoc@Acc & TargetClass@Acc \\
\midrule

Random Select  & - & - & 50.00 & 25.00 & 18.90 & 50.00 \\
InternVL-2.5~\cite{chen2024expanding} & 2B  & 22.37 & 63.34 & 74.45 & 13.33 & 39.79\\
InternVL-2.5~\cite{chen2024expanding} & 8B  & 35.58 & 73.94 & \bf 81.80 & 17.38 & 57.28 \\
LLaVA-NeXT-Video~\cite{liu2024llava,zhang2024llava} & 7B  & 30.63 & 71.92 & 34.08 & 5.39 & 59.90 \\
Qwen2-VL~\cite{wang2024qwen2} & 7B  & 60.18 & 74.15 & 67.18 & 9.58 & 59.19 \\
Qwen2.5-VL~\cite{bai2025qwen2} & 72B  & \textbf{69.97} & \textbf{75.69} & 54.82 & \textbf{29.37} & \bf 61.38 \\

\midrule
Claude 3.7 Sonnet~\cite{claude3_family} & -  & \bf 63.27 & 69.64 & 61.73 & \bf 25.48 & 64.20  \\
GPT-4o-mini~\cite{achiam2023gpt} & -  & 53.40 & \bf 73.67 & \bf 68.52 & 21.29 & \bf 65.12 \\
\bottomrule
\end{tabular}
}
\vspace{-0.1cm}
\caption{Experiments on zero-shot performance of SOTA VLMs, including closed sourced multimodal LLMs for Social Gesture VQA}
\label{tab:vlms}
\vspace{-0.2cm}
\end{table*}

\begin{table*}[!t]
\renewcommand{\arraystretch}{1.1}
\renewcommand{\tabcolsep}{1.6mm}
\centering
\adjustbox{max width=\linewidth}{
\begin{tabular}{lccccccc}
\toprule
\multirow{2}{*}{\textbf{Model}}  & \multirow{2}{*}{\textbf{Input}} & \textbf{Global Perception (\%)} & \multicolumn{2}{c}{\textbf{Gesture Understanding (\%)}} & \multicolumn{3}{c}{\textbf{Gesture Localization (\%)}} \\ 
\cmidrule(lr){3-3} \cmidrule(lr){4-5} \cmidrule(lr){6-7}
& & HumanCount@Acc & GestureDet@Acc & GestureClass@Acc & TargetLoc@Acc & TargetClass@Acc \\
\midrule
Qwen2-VL-7B & {video} & 60.18 & 74.15 & 67.18 & 9.58 & 59.19 \\
Qwen2-VL-7B + LoRA~\cite{hu2022lora} SFT & {video} & 82.55 & 33.07 & {84.52} & 49.21 & 41.19 \\
Qwen2-VL-7B + Full SFT & {video} & {83.64} & 33.07 & 81.13 & {51.25} & 38.51 \\
\bottomrule
\end{tabular}
}
\vspace{-0.1cm}
\caption{Experiments on SFT performance of SOTA VLMs, including closed sourced multimodal LLMs for Social Gesture VQA}
\label{tab:vlms_sft}
\vspace{-0.3cm}
\end{table*}

\subsection{SocialGesture VQA}

From the results of previous tasks, we found that distinguishing social gestures is still very challenging for current SOTA methods in gesture localization and recognition. This difficulty highlights the unique nature of social gesture, where models need to have a unified view of multi-person social interactions and can distinguish details that are hard to detect. This is something traditional video analysis models still struggle with~\cite{abbas2018video,fu2024video}. Inspired by recent progress in multimodal video grounding and video-based VQA~\cite{mu2024snag,chen2024grounded}, we aim to leverage the reasoning capabilities of the latest VLMs and evaluate their performance on this complex task.

\vspace{0.1cm}
\noindent \textbf{Baselines.} 
To assess performance on the SocialGesture VQA benchmark, we chose several SOTA publicly available VLMs, including Qwen2-VL~\cite{wang2024qwen2}, Qwen2.5-VL~\cite{bai2025qwen2}, InternVL-2.5~\cite{chen2024expanding}, and LLAVA-NEXT-Video (LLaVA 1.6)~\cite{liu2024llava}. Additionally, we included closed-source multimodal LLMs such as GPT-4o-mini~\cite{achiam2023gpt} and Claude-3.7-sonnet~\cite{claude3_family} for a broader comparison. Note that GPT-4o~\cite{achiam2023gpt} was excluded, as it was used to generate some of the VQA instruction-following training data, which could lead to biased results.

\vspace{0.1cm}
\noindent \textbf{Evaluation Metrics.} 
Since all VQA pairs can be transformed into a multi-option output, we use rule-based accuracy metrics to evaluate the VQA tasks. Each model was given sufficient context for the task, followed by a question and multiple answer choices. Accuracy was determined by comparing the model’s response with the correct answer. We evaluated models across all subtasks including global perception, gesture understanding, and gesture localization in Section~\ref{subsec:socialgesturevqa}. The accuracy ($\operatorname{Acc}$) in Table~\ref{tab:vlms} is defined as: $ \operatorname{Acc} = \frac{1}{N}\sum_{k=1}^{N}\operatorname{MATCH}_k$, where $N$ is the total number of VQA samples and $\operatorname{MATCH}_k$ is a binary value. If the option in the model output matches the ground truth option, it is 1, otherwise it is 0.

\vspace{-0.3cm}
\paragraph{Results and Findings.}

In Table~\ref{tab:vlms}, the experimental results show that while most VLMs can handle global perception and gesture understanding tasks without finetuning, they all struggle significantly with gesture localization tasks.
This trend is understandable, as localization tasks inherently involve more ambiguity compared to perception and understanding tasks. While there is not a very clear correlation between the number of parameters and performance, Qwen2.5-VL, the largest of our open sourced models, performed well overall.
To explore the potential for improvement, we fine-tuned Qwen2-VL-7B in Table~\ref{tab:vlms_sft}, which led to enhanced performance in most metrics, except those related to gesture binary classification and target identification. Performance drops in some metrics can be attributed to easy overfitting on binary cases. An additional noteworthy observation is that despite being pretrained on vast multimodal datasets, none of the SOTA VLMs achieved over 70\% accuracy on the simplest task: counting humans in social interaction scenes.

These findings indicate that the SocialGesture tasks are significantly more challenging for VLMs than typical VQA benchmarks. This difficulty stems from the natural complexity of real-world social interaction videos used in our benchmark. Unlike controlled environments, our dataset contains realistic scenarios where people frequently overlap with each other, are partially visible, or extend beyond the camera frame, creating natural occlusion cases. In addition, there exists high interdependence between tasks in SocialGesture VQA, which makes it challenging. For instance, when models process such videos, they should first assess the overall scene despite these visual challenges, counting people and understanding actions. This process involves high-level reasoning and the ability to sequentially infer relationships—skills that are currently lacking in SOTA VLMs. Developing models capable of such contextual social visual reasoning remains a key challenge for future research.



\section{Conclusion}

We introduce SocialGesture, the first dataset and benchmark specifically designed for understanding and analyzing multi-person social gestures. We provide a comprehensive definition of tasks related to social gesture recognition, including a novel VQA dataset that encompasses global perception, gesture understanding, and spatial localization. To advance the field of human gesture and social interaction understanding, we benchmarked a variety of current state-of-the-art methods, including action recognition models and vision-language models (VLMs). Our findings reveal that multi-person social gesture understanding requires incremental visual reasoning—area where existing computer vision models fall short. This work offers insights into the complexities of multi-person social interactions and highlights the need for continued research into developing models capable of sophisticated visual reasoning. We hope that SocialGesture will motivate further innovations in gesture understanding.

\section*{Acknowledgment}

We are thankful to the \href{https://keymakr.com/}{Keymakr} for their great work in data annotation. This work was supported in part by the NRF grant funded by the Korea government(MSIT) (No. RS-2025-00563942).

{
    \small
    \bibliographystyle{ieeenat_fullname}
    \bibliography{main}
}

\clearpage

\clearpage
\setcounter{page}{1}
\maketitlesupplementary

\section{Broader Impacts} 
\label{sec:broader} 

We highlight the following social impacts of our work:

\begin{enumerate} 

\item \textit{\textbf{First Multi-person Gesture Dataset:}} By introducing multi-person scenes into gesture recognition, our work can lead to the development of more robust social agents. These models are expected to perform more reliably across diverse real-world scenarios, benefiting human-AI interaction applications such as online education and collaborative virtual environments.

\item \textit{\textbf{Bringing Gesture Recognition to the Large Language Model Era:}} We provide a new Visual Question Answering (VQA) setting to evaluate the performance of Vision-Language Models (VLMs) and multimodal Large Language Models (LLMs) in human gesture understanding. Experimental results reveal that gesture understanding remains a challenging task for these models, highlighting areas for future research and improvement.

\end{enumerate}

\section{Additional Information for SocialGesture} 

We leverage two main data resource: YouTube and Ego4D. The data collections and annotations have been approved by the Institutional Review Board (IRB). Ego4D dataset contains the games of One Night Ultimate Werewolf and The Resistance: Avalon. To fullfill the \href{https://support.google.com/youtube/answer/9783148}{Fair Use Policy} of public YouTube data, we choose to use Creative Commons Attribution Non Commercial 3.0 (cc-by-nc-3.0) to release the annotations. This will provide free and complete access to the research community, excluding only commercial use. 

12 annotators (4 male, 8 female) labeled the SocialGesture dataset, all of whom held at least a bachelor’s degree. The annotators had an average age of 39 years (ranging from 22 to 56). We employed a structured three-step annotation methodology consisting of annotation, verification, and quality assurance. Annotation consistency was evaluated using inter-annotator agreement and Cohen’s kappa. For ambiguous cases and final decisions, we adopted a three-person consensus approach to ensure accuracy and reliability.

\begin{figure*}[htbp]
    \centering
    \begin{minipage}[t]{0.24\linewidth}
        \begin{minipage}{0.95\textwidth}
            \centering
            \scriptsize
            \includegraphics[width=\textwidth]{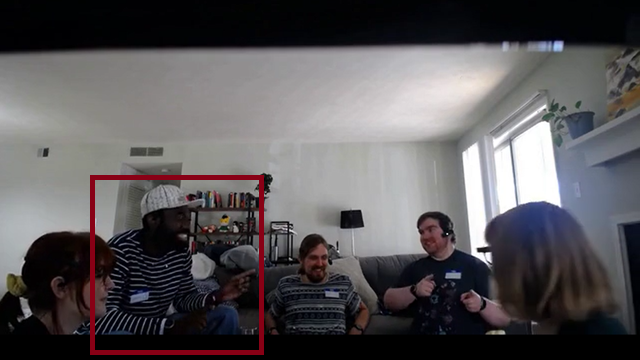}
        \end{minipage}
        \begin{minipage}{0.95\textwidth}
            \centering
            \scriptsize
            \includegraphics[width=\textwidth]{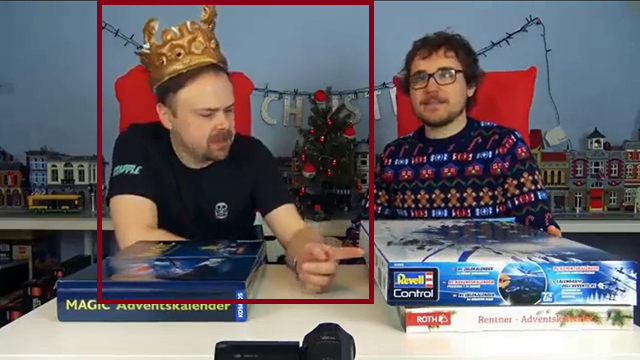}
        \end{minipage}
        \begin{minipage}{0.95\textwidth}
            \centering
            \scriptsize
            \includegraphics[width=\textwidth]{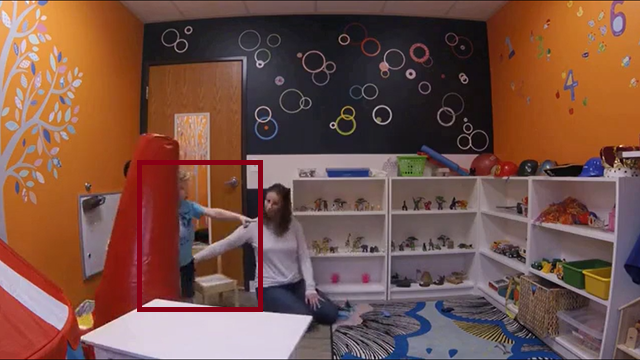}
        \end{minipage}
        \begin{minipage}{0.95\textwidth}
            \centering
            \scriptsize
            \includegraphics[width=\textwidth]{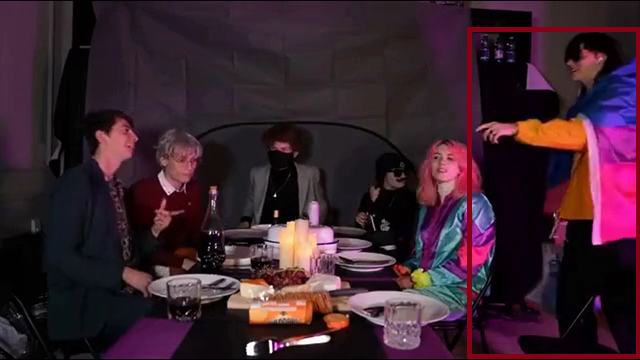}
        \end{minipage}
        \begin{minipage}{0.95\textwidth}
            \centering
            \scriptsize
            \includegraphics[width=\textwidth]{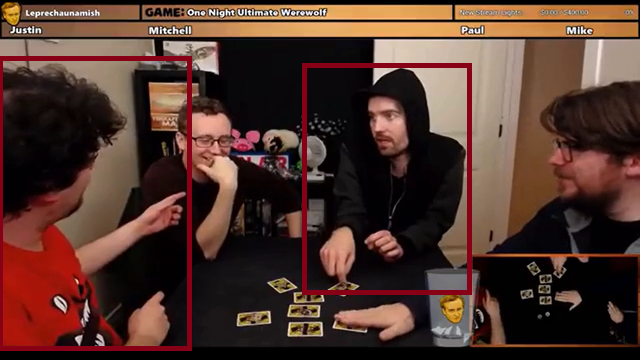}
        \end{minipage}
        \caption*{Pointing Gesture}
    \end{minipage}
    \begin{minipage}[t]{0.24\linewidth}
        \begin{minipage}{0.95\textwidth}
            \centering
            \scriptsize
            \includegraphics[width=\textwidth]{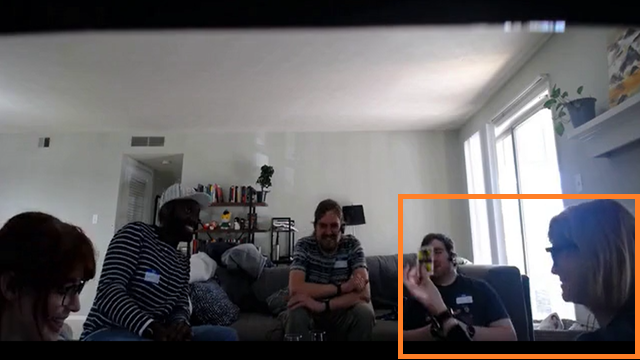}
        \end{minipage}
        \begin{minipage}{0.95\textwidth}
            \centering
            \scriptsize
            \includegraphics[width=\textwidth]{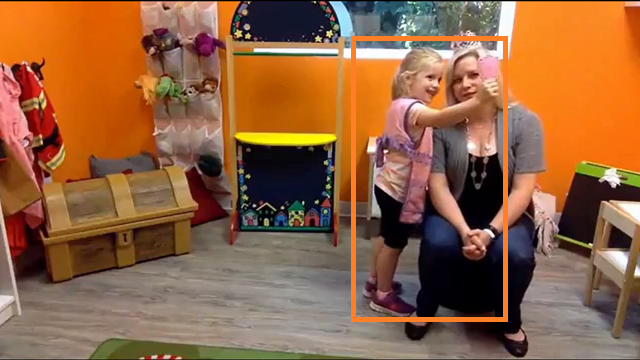}
        \end{minipage}
        \begin{minipage}{0.95\textwidth}
            \centering
            \scriptsize
            \includegraphics[width=\textwidth]{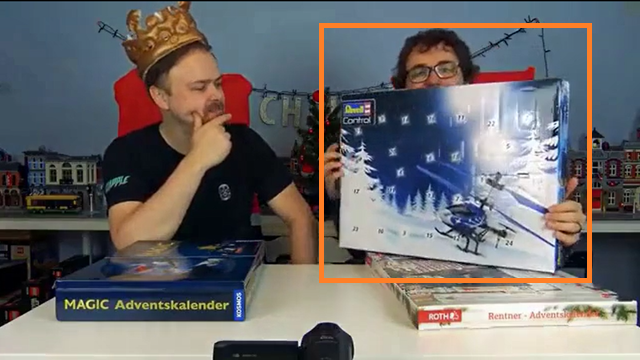}
        \end{minipage}
        \begin{minipage}{0.95\textwidth}
            \centering
            \scriptsize
            \includegraphics[width=\textwidth]{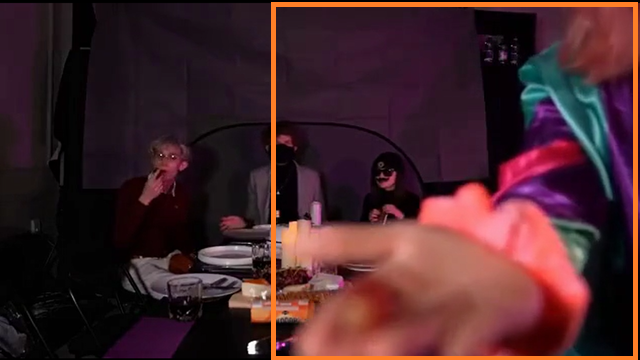}
        \end{minipage}
        \begin{minipage}{0.95\textwidth}
            \centering
            \scriptsize
            \includegraphics[width=\textwidth]{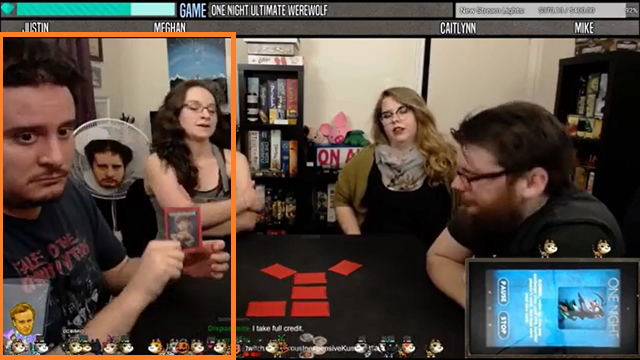}
        \end{minipage}
        \caption*{Showing Gesture}
    \end{minipage}
    \begin{minipage}[t]{0.24\linewidth}
        \begin{minipage}{0.95\textwidth}
            \centering
            \scriptsize
            \includegraphics[width=\textwidth]{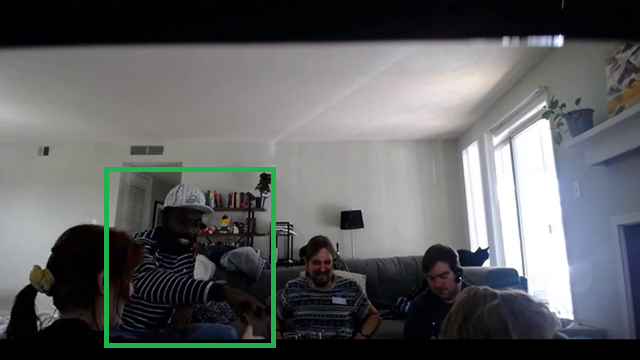}
        \end{minipage}
        \begin{minipage}{0.95\textwidth}
            \centering
            \scriptsize
            \includegraphics[width=\textwidth]{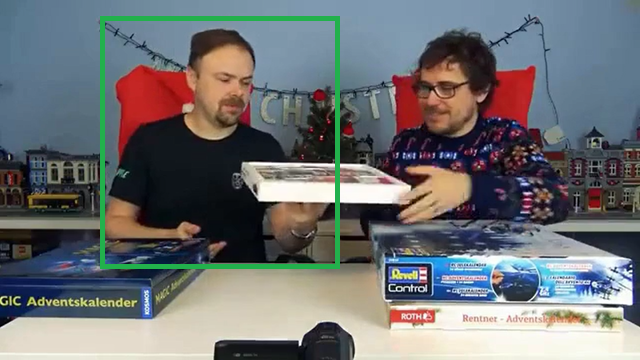}
        \end{minipage}
        \begin{minipage}{0.95\textwidth}
            \centering
            \scriptsize
            \includegraphics[width=\textwidth]{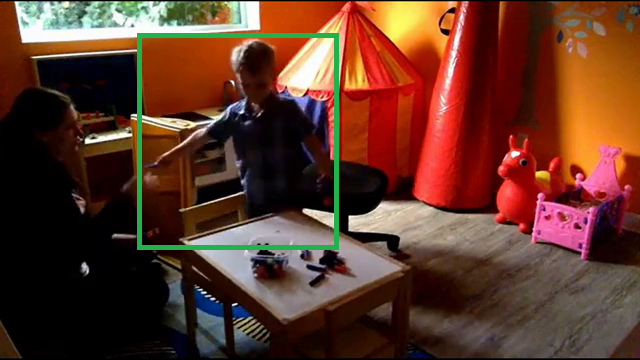}
        \end{minipage}
        \begin{minipage}{0.95\textwidth}
            \centering
            \scriptsize
            \includegraphics[width=\textwidth]{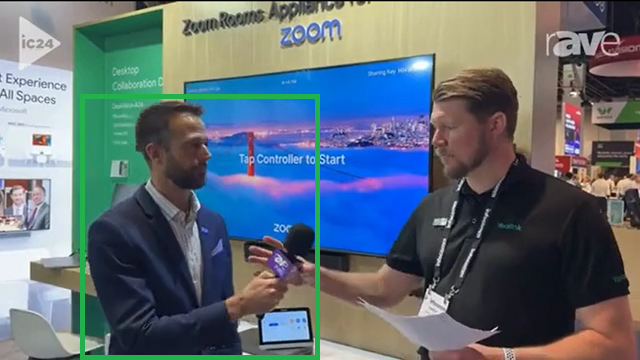}
        \end{minipage}
        \begin{minipage}{0.95\textwidth}
            \centering
            \scriptsize
            \includegraphics[width=\textwidth]{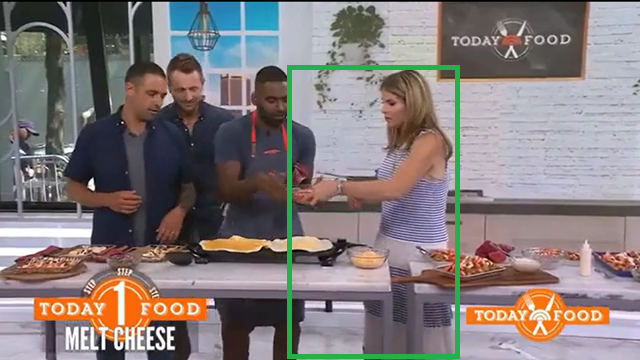}
        \end{minipage}
        \caption*{Giving Gesture}
    \end{minipage}
    \begin{minipage}[t]{0.24\linewidth}
        \begin{minipage}{0.95\textwidth}
            \centering
            \scriptsize
            \includegraphics[width=\textwidth]{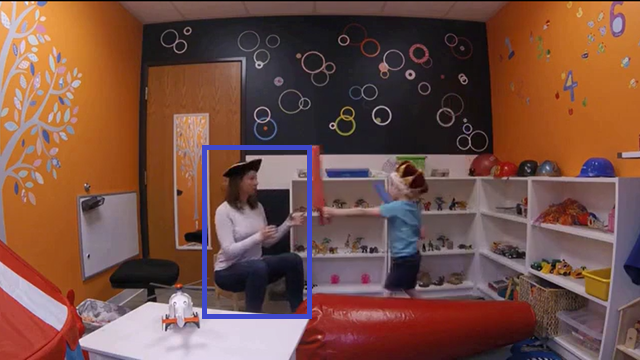}
        \end{minipage}
        \begin{minipage}{0.95\textwidth}
            \centering
            \scriptsize
            \includegraphics[width=\textwidth]{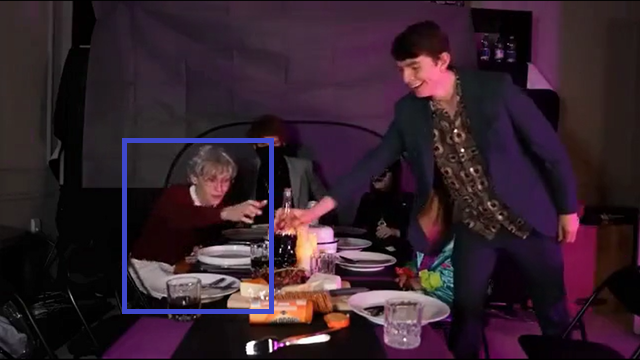}
        \end{minipage}
        \begin{minipage}{0.95\textwidth}
            \centering
            \scriptsize
            \includegraphics[width=\textwidth]{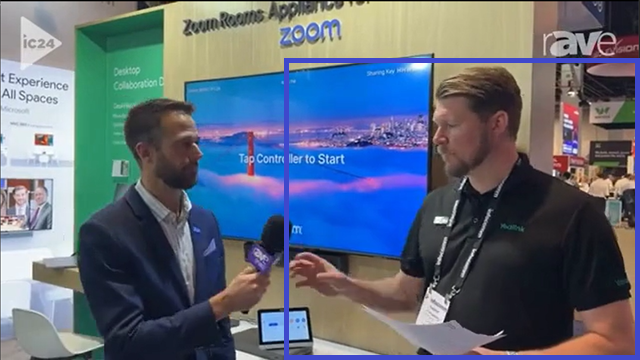}
        \end{minipage}
        \begin{minipage}{0.95\textwidth}
            \centering
            \scriptsize
            \includegraphics[width=\textwidth]{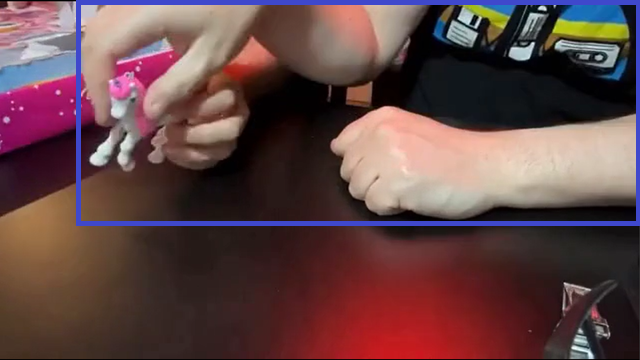}
        \end{minipage}
        \begin{minipage}{0.95\textwidth}
            \centering
            \scriptsize
            \includegraphics[width=\textwidth]{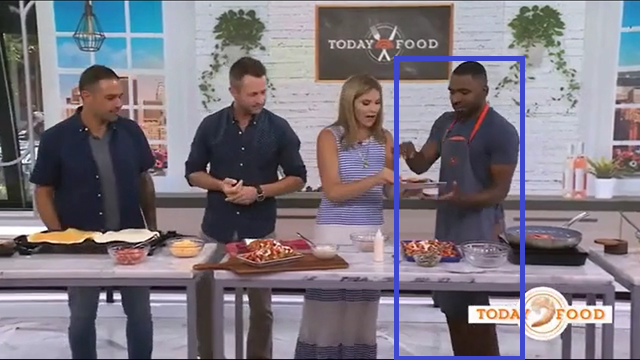}
        \end{minipage}
        \caption*{Reaching Gesture}
    \end{minipage}
\caption{The diversity of four social gesture in our dataset.}
\label{fig:appendix:socialgesture_example}
\end{figure*}

As shown in Table~\ref{tab:dataset_comparison}, our dataset is diverse and includes many different social scenes. Figure~\ref{fig:appendix:socialgesture_example} presents examples of social gestures in various settings. In {SocialGesture}, we observe that the pointing gesture constitutes a higher proportion compared to other gestures. The frequencies of showing, giving, and reaching gestures are approximately equal.

Moreover, the manifestation and proportion of gestures vary significantly across different scenes. For instance, in social game scenes such as \textit{One Night Ultimate Werewolf}, the majority of social gestures are pointing gestures. This is because \textit{One Night Ultimate Werewolf} is a game where participants need to vote for the werewolf during gameplay, making pointing gestures highly prevalent. In contrast, giving and reaching gestures have a higher proportion in collaborative cooking videos and interview settings. All four gesture types occur with similar frequency in children education videos. This is because gestures play a crucial role for children in expressing their unspoken thoughts to listeners, as young children have not yet fully developed strong spoken language abilities. 

\section{Discussion} 
\label{sec:limitation} 

\begin{enumerate} \item \textit{\textbf{Imbalanced Distribution of Social Gestures:}} 

Since the videos in our dataset are from natural scenes, the distribution of social gesture is imbalanced. The proportions of gesture types in our dataset are (pointing:reaching:giving:showing = 13.03:1.23:1.22:1.00). It reflects the natural distribution of gestures in context. Standard datasets, such as MS-COCO, also exhibit such imbalances (e.g., 20 times more people than bicycles), which are an inherent property of natural scenes. Our dataset is also similarly naturally imbalanced, e.g. the gestures reaching, giving, and showing are rarer than pointing because they require object interactions. We address imbalance in our multi-class classification experiments by creating a balanced subset. 


\item \textit{\textbf{Granularity of Categorization:}} 

While we have categorized social gestures into four types based on David McNeill's theory of deictic gestures, this classification is coarse considering the wide range of such gestures. It's possible that within each category, there are more subtle, fine-grained gesture types not explicitly accounted for. For example, the showing gesture can be divided into showing to a person on-screen and showing to an off-screen audience. Similar distinctions can occur based on different video perspectives, such as egocentric (first-person) or exocentric (third-person) views. Future work could involve developing more granular classifications and corresponding evaluation metrics to provide a deeper understanding of social gestures. 

\item \textit{\textbf{Multi-person Gesture and Gaze:}} Both gaze and gesture serve as crucial non-verbal social cues and share several common characteristics. Notably, recent Transformer-based approaches for gaze target estimation ~\cite{ryan2025gaze,song2024vitgaze} have potential to be adapted for detecting the targets of pointing gesture. To facilitate further research, we will release the pointing gesture subset of SocialGesture, providing a benchmark for these methods and enabling exploration of co-occurrence scenarios of gaze and pointing gestures.

\end{enumerate}

\end{document}